\documentclass{article} % For LaTeX2e
\usepackage{iclr2025_conference,times}

\usepackage{amsmath,amsfonts,bm}

\def\eqref#1{equation~\ref{#1}}
\def\1{\bm{1}}

\DeclareMathAlphabet{\mathsfit}{\encodingdefault}{\sfdefault}{m}{sl}
\SetMathAlphabet{\mathsfit}{bold}{\encodingdefault}{\sfdefault}{bx}{n}

\usepackage{hyperref}
\usepackage{url}

\usepackage{ifthen}
\usepackage{xcolor}
\usepackage{xspace}

\usepackage{enumitem}

\usepackage{subcaption}
\usepackage{array}
\usepackage{placeins}

\usepackage{graphicx}
\usepackage{tikz}
\usepackage{wrapfig}
\usetikzlibrary{arrows.meta,calc,shapes.geometric}
\usepackage{enumitem}

\usepackage{cleveref}

\crefname{figure}{fig.}{figs.}
\Crefname{figure}{Fig.}{Figs.}
\crefformat{section}{\S#2#1#3}
\crefmultiformat{section}{\S#2#1#3}{ and \S#2#1#3}{, \S#2#1#3}{, and\S#2#1#3}
\Crefformat{section}{\S#2#1#3}
\crefformat{subsection}{\S#2#1#3}
\Crefformat{subsection}{\S#2#1#3}
\crefmultiformat{subsection}{\S#2#1#3}{ and \S#2#1#3}{, \S#2#1#3}{, and\S#2#1#3}
\crefformat{subsubsection}{\S#2#1#3}
\Crefformat{subsubsection}{\S#2#1#3}
\crefmultiformat{subsubsection}{\S#2#1#3}{ and \S#2#1#3}{, \S#2#1#3}{, and\S#2#1#3}

\title{AgSpec: Pushing the Limits of Retrieval-Based Speculative Decoding in Coding Agent Pipelines}

\iclrfinalcopy

\author{Sumin Lee$^{*}$,
Sukmin Cho$^{*}$,
Seungjae Lim,
Youngjin Kwon$^{\dagger}$ \\
School of Computing, KAIST \\
\texttt{sum37@kaist.ac.kr}, \texttt{\{smcho,sjlim,yjkwon\}@casys.kaist.ac.kr} \\
$^{*}$Equal contribution.
\quad
$^{\dagger}$Corresponding author.
}

\newcommand{\sys}{\textsc{AgSpec}\xspace}

\newboolean{publicversion}
\setboolean{publicversion}{false}     % draft: show comments
\ifthenelse{\boolean{publicversion}}
  {\newcommand{\grumbler}[3]{}}
  {\newcommand{\grumbler}[3]{\textcolor{#1}{\textbf{#2: #3}}}}

\ifthenelse{\boolean{publicversion}}
  {
   \newcommand{\tbdnote}[1]{\rule{4em}{1.6ex}}}
  {
   \newcommand{\tbdnote}[1]{\textcolor{red}{\textbf{[TBD: #1]}}}}

\begin{document}

\maketitle
\lhead{Preprint}

\begin{abstract}
Retrieval-based speculative decoding (SD) drafts tokens by copying continuations
from existing text, which suits coding agents that repeatedly reproduce code, logs,
and earlier attempts. Yet existing methods fall short in agent pipelines: much of
the reusable text is missing from their corpora or stored in a form that differs
from what the agent emits, and their draft lengths ignore that accept length varies
across agents and drifts over turns. We present \sys, a framework that supplies the corpus and draft-length policies that existing retrieval engines lack in coding-agent pipelines. \sys retrieves from
session, workspace, and global corpora, retaining the ongoing session trajectory
and indexing opened files in the agent's emission format. It bounds each agent's
draft length with an offline-profiled cap and adapts the length online from
verification feedback. On two repository-level multi-agent coding benchmarks, \sys outperforms
five retrieval-based drafters and EAGLE-3 in most evaluated settings,
raising generation throughput over autoregressive decoding up to
4.37$\times$ at batch size 1 and 4.76$\times$ at batch size 16.
\sys also remains effective on benchmarks without a repository or a
multi-agent pipeline, showing that its gains generalize to coding agents broadly.
\end{abstract}

% Section structure. Each file opens with a "% SPINE:" comment stating that section's
% claim in one line, and carries its own \section / \subsection headers.
%
%   2RelatedWork  coding-agent workflows and retrieval-based speculative decoding
%   4Design       Section 3: motivation and design for each drafting decision
%   5Evaluation   Section 4: protocols, published-method comparisons, and ablations
%
% Background leads directly into Design. Each design motivation connects prior
% rationale, a measured observation or pending hypothesis, and the proposed mechanism.
% Published-method workload breakdowns are included within Evaluation.
% Unfilled measurements remain \tbd or explicitly unmeasured red xx placeholders.
% Design was promoted from tex/4Design_outline.tex on 2026-09-15.
% tex/Design.tex and tex/Analysis.tex are reference-only and are not included.
% The TikZ overview is temporary: replace with a PowerPoint drawing or remove
% before submission. See figure/design_overview.tex.
% TODO(check): align the older Intro/Evaluation/Discussion/Conclusion/Appendix
% controller descriptions and result pointers with the new Design (STORYLINE 5.0-B).
%
% HISTORICAL: tex/3Cost.tex was removed on 2026-09-11. Recover it with
% git show b6fc78a:tex/3Cost.tex under a different name.
% tex/3Observation.tex preserves the former standalone section as a reference.
\section{Introduction}
\label{sec:intro}

Coding agents have become a representative LLM application \citep{liu2026claudecode}.
Unlike a chatbot, a coding agent handles a single user request as a multi-turn session that alternates between model calls and tool execution \citep{yang2024sweagent}.
Within a session, the work is often divided among specialized agents, such as a coder that edits files and an executor that runs tests and scripts to check the result \citep{hong2024metagpt}.
As user requests have shifted from synthesizing a single function \citep{chen2021humaneval} to resolving repository-level issues \citep{jimenez2024swebench}, agents rarely succeed in one pass and instead refine their attempts over multiple turns \citep{madaan2023selfrefine}.
Each additional turn can improve the output, but it also lengthens the time a user waits for the result \citep{tiwari2026cachewise}.

Speculative decoding (SD) attacks this cost directly, reducing decoding latency
without changing the output \citep{chen2023accelerating}. In SD, a
drafter proposes several tokens, and the target model verifies them in parallel and commits the accepted tokens.
Each accepted token saves a sequential target-model step.
This saving comes at the cost of generating draft tokens and verifying tokens that are later rejected.
Rejected tokens are nearly free at small batch sizes but become expensive at large batch sizes, where verification demands much more compute \citep{liu2024smartspec,sadhukhan2024magicdec}.
SD thus pays off only when the drafter is both cheap and accurate. Existing
designs range from small draft models \citep{leviathan2023fast} and trained draft
heads \citep{li2025eagle3} to retrieval-based SD, which retrieves draft tokens
from a text corpus.

Retrieval-based SD is particularly effective for code generation \citep{liu2026illusion}. 
It copies continuations from text the system already has, such as the current prompt, earlier responses, or a prebuilt datastore, without running a separate draft model. 
Code suits this copying, since identifiers, code fragments, and repository conventions recur, and coding-agent sessions offer still more of it, as repeated subtasks and refinement loops produce long, predictable sequences. 
However, previous works do not elicit the full performance of retrieval-based SD at coding agents.
They overlook what makes a coding agent different, and leave much of that reuse out of reach.

\begin{figure}[!t]
\centering
\includegraphics[width=\linewidth]{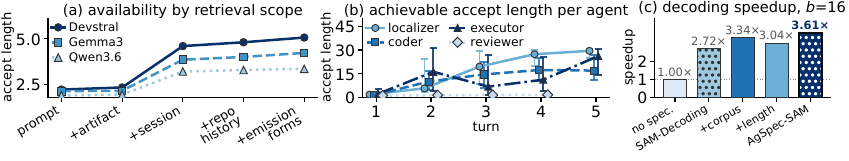}
\caption{
\textbf{(a)} Accept length when restoring availability, then
matchability. 
\textbf{(b)} Achievable accept length for Gemma3-27B, taking
the best available continuation at each step. Markers are medians and bars the interquartile range.
\textbf{(c)} Decoding throughput of SAM-Decoding and \sys-SAM relative to no
speculation for Gemma3-27B. All data is from SWE-bench Verified.
}
\label{fig:motiv-headroom}
\end{figure}

A draft helps only if two conditions hold: the text the agent reuses must be \emph{available} in the corpus, and it must be \emph{matchable} against what the agent generates.
Existing corpora fall short on both.
First, they limit availability: they hold either a prebuilt datastore or the text
of the current call, whereas coding agents reuse text from the ongoing session and the repository. A file drops out of
reach once the prompt truncates it, and a retry cannot draw on the previous
attempt once its call ends; covering these requires corpora that span the session
and the repository.
Second, they limit matchability: they store text as the agent read it, yet a coder
emits a patch whose copied lines each carry a diff prefix absent from the file.
\Cref{fig:motiv-headroom}(a) shows that expanding retrieval scope raises accept length on every model, with the largest gain from session history and smaller additional headroom from repository history and emission-form indexing.

Even when a reusable match exists, choosing the draft length remains critical, as
it directly determines how many drafted tokens are rejected.
Existing methods fix the draft length policy before a session runs, deriving it from
a static cap \citep{saxena2023prompt,samdecoding2025,he2024rest,zhao2025fastcoder}
or from properties of the match such as its length \citep{suffixdecoding2025}
or occurrence count \citep{he2024rest,zhao2025fastcoder,suffixdecoding2025}.
However, in coding agents, the achievable accept length follows the agent pipeline.
\Cref{fig:motiv-headroom}(b) shows that each agent occupies a distinct accept-length range that shifts differently across and within turns.
Each agent and context thus has a characteristic accept length, which can guide
the draft length better than a fixed or match-derived policy.

These limitations stem not from how existing methods match text but from the
policies around the matching. A retrieval-based SD method pairs a retrieval
engine that finds a continuation for the current context (\textit{e.g.,} a
suffix tree in SuffixDecoding \citep{suffixdecoding2025}) with policies that
decide what the corpus holds and how many tokens to draft. Existing policies
are built around the individual request, the unit an inference engine serves:
they index the current call, a prebuilt datastore, or outputs pooled across
requests \citep{saxena2023prompt,he2024rest,suffixdecoding2025}, and size
drafts by a static cap or the match itself. In a coding agent, however, reuse
and acceptance depend on structure above the request, namely the session, the
repository, and the role of each agent. 

We present \sys, a retrieval-based SD
framework that builds its policies on this structure while leaving the engine
unchanged, so it applies to existing engines as is.
\sys realizes each policy with a simple yet effective design.
\textbf{(1) Coding-agent-aware retrieval.} For the corpus, \sys organizes text
by origin into session, workspace, and global reference corpora. The workspace
corpus indexes the artifacts a session opens, in full and in the forms the
agent emits. \textbf{(2) Agent-aware draft-length control.} For the draft
length, \sys profiles agent-specific accept lengths offline to set per-agent
draft caps, and learns a draft-length scale that adapts the length per turn
and per request from verification feedback.

% To these, we present \sys, a retrieval-based SD framework tailored to coding-agent pipelines. 
% \sys addresses each of these problems with a simple yet effective solution. \textbf{(1) Coding-agent-aware retrieval.} \sys organizes text by origin into session, workspace, and global reference corpora. The workspace corpus indexes the artifacts a session opens, in full and in the forms the agent emits. \textbf{(2) Agent-aware draft-length control.} \sys profiles agent-specific accept lengths offline to set per-agent draft caps, and learns a draft-length scale to adapt the proper length per turn and per request from verification feedback. 
% Both designs of \sys sit above the existing retrieval engine, so \sys extends existing methods without modification.

%\yj{인트로를 읽고 난 후에도 SuffixDecoding과 position 불분명합니다. Figure 4에서 suffixdecoding보다 throughput 17\% 향상되었는데, 가장 근본적인 이유가 무엇인가요?
%Framework으로서 어떠한 부분을 향상시켰나요?}
%\smc{Design 설명을 두 문단으로 나누어서, 앞선 문단에서는 Retrieval engine과 policy에 대해 설명하고, ours는 coding agent에 적합한 policy를 제공하는 프레임워크라고 설명을 추가하였습니다. plot은 아직 나오는 중입니다.}

We compare \sys against AR decoding, five retrieval-based SD methods
\citep{saxena2023prompt,he2024rest,suffixdecoding2025,samdecoding2025,zhao2025fastcoder},
and EAGLE-3 \citep{li2025eagle3} on two repository-level benchmarks, SWE-bench
and TeamBench \citep{jimenez2024swebench,teambench2026}. \sys consistently outperforms most baselines.
Especially, with SAM-Decoding as its retrieval engine, \sys raises the
speedup of Gemma3-27B on SWE-bench over AR decoding from 2.72$\times$
to 3.61$\times$ at batch size 16 (Fig.~4).
\sys also generalizes beyond this setting, achieving the highest throughput
on single-agent and non-repository benchmarks.
These results highlight the effectiveness of \sys as a training-free
acceleration framework for coding agent applications.

\section{Background}
\label{sec:related}
\label{sec:background}

\subsection{Coding Agents}
\label{sec:agents}

Coding agents resolve software engineering tasks by handling each user request as
a \emph{session} \citep{jimenez2024swebench}.
A session may span several agents, each of which alternates between
\emph{model calls} and tool executions.
Throughout the session, agents operate on a persistent \emph{workspace}: the
repository checkout together with its execution environment.
The workspace holds the \emph{artifacts} agents work on, such as repository
files, patches, and tool outputs.
Unlike conversation history, artifacts are workspace state; in particular, most
repository files predate the session and outlive it.
However, what the workspace holds differs from what a model call sees: each
prompt exposes only the artifacts the scaffold selects, often truncated to fit the
context length \citep{yang2024sweagent, wang2025openhands}.

\begin{wrapfigure}[18]{r}{0.31\textwidth}
    \vspace{-\intextsep}
    \centering
    \includegraphics[width=\linewidth]{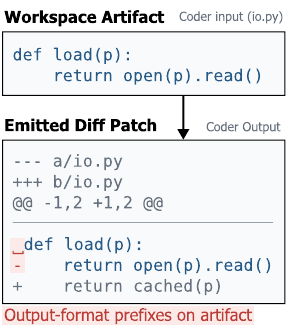}
    \vspace{-18pt}
    \caption{The same code in two forms: as the workspace holds it, and as a
    unified-diff patch emits it. Red marks the prefix each copied line gains.}
    \label{fig:emission-form}
\end{wrapfigure}

Even when an agent reuses workspace text, its output rarely reproduces it
verbatim. An agent reads an artifact as it stands but writes it back in the
scaffold's \emph{emission format}.
Under unified-diff editing \citep{gnudiffutils}, a call emits a header naming the
file, followed by kept lines prefixed with a space and removed lines prefixed with
a minus, so every line copied from the artifact begins with a character the
artifact does not contain (\Cref{fig:emission-form}).
The code is identical, but the token sequence differs, so text indexed as the
agent read it may not match what the agent generates.
Other scaffolds emit whole files, search-and-replace blocks, or line-addressed
edit commands \citep{xia2025agentless,wang2025openhands,yang2024sweagent}, and the
choice measurably affects editing quality \citep{gauthier2023udiff}.
Because the format is a property of the scaffold, it is known before the session
runs.

The extent to which a call's output repeats earlier text also depends on what the call does and when it is made.
Within a session, each model call is issued by an agent, serves a \emph{role}, and belongs to a
\emph{turn}.
Roles are the functions a session performs, and each generates a different kind
of text: a localizer names paths, a coder rewrites files the session has already
read, and an executor reproduces test logs.
A role describes what a call does rather than which agent issues it:
multi-agent systems assign each role to a specialized agent
\citep{hong2024metagpt,qian2024chatdev,hyperagent2024}, whereas
others fix roles as pipeline stages \citep{xia2025agentless} or interleave them
within one agent's loop \citep{yang2024sweagent}.
A turn comprises a solution attempt and its execution feedback, possibly spanning
several roles; failed attempts trigger further turns until the task completes or a
retry limit is reached \citep{yang2024sweagent}.
Across turns, agents return to the same files and repeat much of the previous
attempt.
Roles thus determine what kind of text a call generates, and turns determine how
much of it repeats.

\subsection{Retrieval-Based Speculative Decoding}
\label{sec:prelim}

Retrieval-based SD obtains drafts from existing text rather
than a draft model: it matches a suffix of the decoding context, such as the prompt
and generated prefix, against a corpus and copies the continuation that follows.
Its effectiveness hinges on two design choices.
The \emph{corpus scope}, which text is indexed and how long it is retained,
determines whether a useful match exists.
The \emph{draft length}, how many matched tokens are proposed, determines how many
of them are rejected.

Existing methods differ in both what their corpora index and how long they retain
it. Request-local corpora last for a single call. PLD searches the current prompt
and generated prefix \citep{saxena2023prompt}, and SAM-Decoding and SuffixDecoding
build a per-request suffix automaton or tree over the prompt and output
\citep{samdecoding2025,suffixdecoding2025}. 
Prebuilt corpora are assembled before
serving and stay fixed. REST retrieves from a text or code datastore \citep{he2024rest}, SAM-Decoding pairs its request-local automaton with a static
datastore \citep{samdecoding2025}, and FastCoder adds repository-local files to
common code \citep{zhao2025fastcoder}. A few corpora persist across requests:
SuffixDecoding keeps a global tree of past responses
\citep{suffixdecoding2025,arcticSuffixCache2026}, and FastCoder caches verified
and generated sequences \citep{zhao2025fastcoder}.

Draft lengths are either fixed or scaled by the match. PLD and SAM-Decoding cap
each continuation at a fixed length, and REST uses a fixed tree length
\citep{saxena2023prompt,samdecoding2025,he2024rest}. SuffixDecoding scales its
length with match length by a configured factor \citep{suffixdecoding2025}, so
its draft length varies with the input, but the factor itself is set before
serving.

These methods accelerate both code generation and agent workloads
\citep{he2024rest,zhao2025fastcoder,suffixdecoding2025}, yet they remain limited
in the agentic setting. First, agents reuse text from both the ongoing session and the repository workspace, and emit it in the scaffold’s emission format. A corpus should therefore span both 
sources and index text in that format, yet existing methods each cover only some
of these sources. Second, their draft lengths ignore the contexts that shape
acceptance: they are fixed \citep{saxena2023prompt, he2024rest, samdecoding2025}
or scaled by a preset factor \citep{suffixdecoding2025}, never corrected by
verification outcomes, and blind to roles and turns. \sys addresses these
limitations with a source- and emission-aware corpus
(\Cref{sec:design-retrieval}) and an adaptive draft
length (\Cref{sec:design-length}). 
% \sjl{라임 맞춰놨슴다}
% \input{tex/3Observation}

\section{\sys Design}
\label{sec:design}

\begin{figure}[t]
\centering
\includegraphics[width=\linewidth]{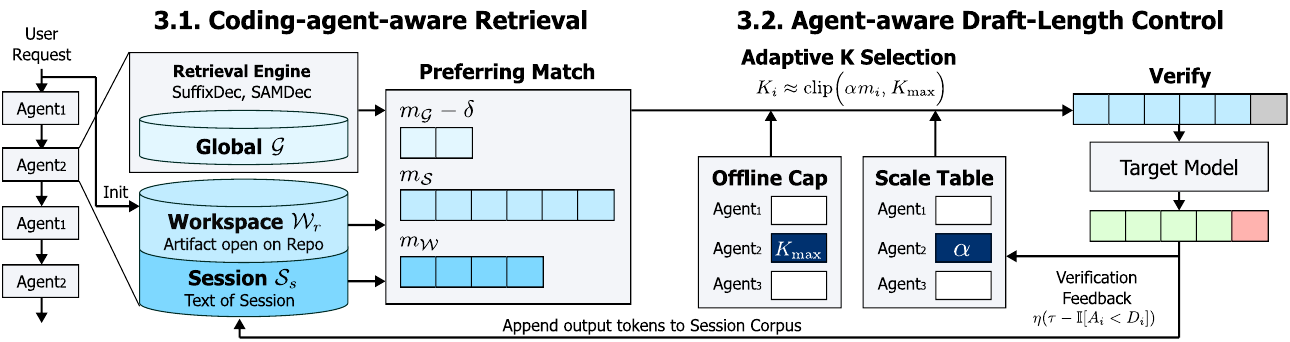}
\vspace{-16pt}
\caption{Overview of \sys. Retrieval consults the session, workspace, and
global corpora. Verification extends the output, updates the session
corpus, and provides feedback to the length controller. The workspace corpus grows as artifacts are opened and is discarded
when the session ends.}
\label{fig:design-overview}
\end{figure}

\sys is a retrieval-based SD framework for coding agents that builds on an existing retrieval engine.
A retrieval engine finds the longest suffix match of the decoding context in a corpus and returns its continuation, using an index such as the suffix automaton in SAM-Decoding \citep{samdecoding2025} or the suffix tree in SuffixDecoding \citep{suffixdecoding2025}.
\sys operates on top of the engine without modifying it, so it applies to any engine that can search multiple corpora and accept a per-step draft length.
\sys combines corpora organized by token source (\Cref{sec:design-retrieval}) with verification-driven draft-length control (\Cref{sec:design-length}). 
% \sjl{이거 아래랑 이름 같게 안해도 될까요}
\Cref{fig:design-overview} depicts the overview of \sys. 
% At each step, \sys queries the retrieval engine over multiple corpora for a continuation, sets the draft budget, and sends the drafted tokens to the target model (\Cref{fig:design-overview}).

% \sys requires only that the engine index multiple corpora separately, report the longest match and its occurrences in each, and accept a per-step draft budget; matching and occurrence ranking remain unchanged.
% Any engine that meets these requirements can host \sys, including both of the above.

% The overview depicts logical dependencies, not a verified concurrent schedule.
% Its editable TikZ source is figure/design_overview.tex; the three corpora
% use the session terminology and the shared frequency rule.
% The imported Overleaf asset figure/design_overview_2.pdf is retained for
% revision; its two-corpus/recency design predates the current specification.

\subsection{Coding-Agent-Aware Retrieval}
\label{sec:design-retrieval}

\sys addresses the corpus gap by indexing the reusable
text in separate corpora keyed by source. Three mechanisms
build on this separation: each corpus captures one source of reusable text, 
% \sjl{이거 보면서 생각했는데, 그냥 한 corpus에 모든 소스 때려넣은 거랑 따로 잘라야 하는 이유랑에 대한 모티가 위에 없었던 것 같슴다. 이때까지 전부 이런데서도 가져올 수 있다 여서, 이게 바로 뒤 문단에 나오긴 하는데, 처음 읽을 때 아 corpus에 이런걸 넣어야겠다 (intro+백그) 정도로 이해되서 뭔가 한 문장 백그나 인트로에 이거 용 문장 넣으면 좋을 것 같아요}
workspace artifacts are indexed in the forms the agent emits, and a preference
order ranks matches across corpora by how closely their source relates to the
active session.

\textbf{Three corpora by source.}\;
\sys maintains one corpus per source of reusable text.
The \emph{session corpus} $\mathcal{S}_s$ holds the text of session $s$, including
prompts, tool outputs, and generated tokens, and is shared across its calls and
agents, so text from earlier calls and turns remains retrievable.
The \emph{workspace corpus} $\mathcal{W}_r$ holds the repository artifacts opened
during the current session and grows as additional artifacts are accessed.
The \emph{global corpus} $\mathcal{G}$ holds a large body of static reference text
that is not specific to the current session; it is assembled before
serving and stays fixed.
These corpora are kept separate because their provenance and lifetime differ:
the session and workspace corpora, which we call \emph{live}, are constructed
within the current session and discarded when it ends, whereas the global corpus
remains fixed across sessions.
% \sys preserves each match's source at retrieval time, which lets it rank matches
% by how closely their source relates to the active session. 
% \sjl{매치의 우선순위를 정하는 내용이 굳이 여기 필요할까 합니다 match preferring으로 옮길까 함다}
% Since the partition is by source rather than content, the corpora may overlap, but
% \sjl{These corpora are kept separate because their provenance and lifetime differ, and \sys preserves this source identity at retrieval time.
% Since the partition is by source rather than content,... 로 이어 가는거 어떨까요? 갑자기 3개 소개하고 스플릿 한 이유 보다 이어가는 문장이 필요할 것 같습니다} 

\textbf{Indexing artifacts in emission form.}\;
Covering the workspace is not enough if its text no longer matches what the agent
generates. The first time a call opens an artifact, \sys adds it to
$\mathcal{W}_r$ in two forms. It adds the file as is, so the file remains
retrievable even after truncation drops it from the prompt. It also adds the file
as the agent would write it: because a unified-diff patch prefixes kept lines
with a space and removed lines with a minus (\Cref{fig:emission-form}), \sys adds
one copy with every line prefixed by a space and another with every line
prefixed by a minus. Any line the coder copies into its patch thus matches one of
these copies. A scaffold that writes whole files through a tool call instead
needs one extra copy: the file as that call encodes it, with its newlines and
quotes escaped. Later calls in
the session reuse these entries instead of reindexing the file.

\textbf{Preferring matches by source.}\;
% With several corpora able to match, \sys must choose one to draft from. 
% \sjl{왜 골라야 하는가? 셋 다 쓰면 안되는가? computing bound 쪽 이야기가 필요할 것 같습니다. 그리고 drafting은 왜 다 해놓고 verifing만 하나를 쓰는가? drafting은 미미하기 때문에...}
The retrieval engine queries each corpus with the same suffix of the decoding context
and reports its longest match length $m_{\mathcal{S}}$, $m_{\mathcal{W}}$, and
$m_{\mathcal{G}}$. Following prior retrieval-based SD
\citep{suffixdecoding2025,samdecoding2025}, \sys treats a longer match as a better
draft source, since more of the recent context agrees with the corpus text.
Between the two live corpora, the longer match wins, with ties going to the
session, whose text reflects the ongoing session. The global corpus must beat the
live match by a margin $\delta$, because its text is plentiful and unrelated to
the repository, so a longer match there is likely coincidental.
Formally, \sys drafts from $\mathcal{G}$ only when
$m_{\mathcal{G}} > \max(m_{\mathcal{S}}, m_{\mathcal{W}}) + \delta$;
otherwise, it drafts from $\mathcal{S}_s$ if $m_{\mathcal{S}} \ge m_{\mathcal{W}}$,
and from $\mathcal{W}_r$ if not.
\Cref{sec:app-pref-margin} includes a $\delta$ sweep study.

\textbf{Continuation selection.}\;
Within the chosen corpus, a match may occur at multiple positions with different
continuations, so \sys must select one occurrence. For both the global corpus
$\mathcal{G}$ and the session corpus $\mathcal{S}_s$, \sys groups occurrences
by their next three tokens and selects the most frequent group, breaking ties
by the most recent occurrence. For the workspace corpus $\mathcal{W}_r$, it
selects the first occurrence. 
\Cref{sec:appendix-selection} includes the detailed analysis.
% \sjl{왜 corpus마다 달라야 하나요? 이유가 없으면 너네 셋팅에서 hyperparameter tuning한거 아니냐 물어볼 것 같습니다}
% \smc{이건 뒤에 데이터 나오고 한 번 생각해봅시다.}
% \sml{TBD 채움}

Together, these mechanisms let \sys draft from text that existing corpora
miss: the session's own outputs and whole workspace files in the agent's
emission format, while preferring the source most relevant to the ongoing session.

\subsection{Agent-aware Draft-Length Control}
\label{sec:design-length}

A draft token saves a decoding step if accepted but wastes verification if
rejected, a waste that grows with batch size as verification becomes
compute-bound. The draft length should therefore track how likely each token is to
be accepted, which existing methods misjudge in two ways. A static cap ignores
that accept length differs across agents and drifts across and within turns
(\Cref{fig:motiv-headroom}(b)). Match-derived rules \citep{suffixdecoding2025}
size the length by the match alone, regardless of which agent emits it or
when. Both over-draft where acceptance is short. \sys instead estimates the
draft length from two sources: an offline-profiled cap for stable differences across
agents, and an online scale that adjusts it to runtime drift from verification
feedback. This cuts rejected tokens at a small cost in accepted ones, raising
throughput most at large batch sizes.
% \smc{해당 디자인의 장점을 간단하게 서술했습니다.}

\textbf{Overview. } At decoding step $i$, \sys sets the draft length for the requesting agent $a_i$ as
\begin{equation}
K_i = \mathrm{clip}\big(
  \mathrm{round}(\alpha^{(a_i)} m_i),\,
  1,\,
  K_{\max}^{(M,a_i)}
\big),
\label{eq:design-length}
\end{equation}
where $m_i$ is match length, $\alpha^{(a_i)}$ is the agent's online scale and $K_{\max}^{(M,a_i)}$ is
its offline cap under target model $M$; the lower bound ensures that every match
yields at least one draft token. The drafter copies up to $K_i$ tokens that follow the match, so the actual
draft length is $D_i = \min\{K_i, \ell_i\}$, where $\ell_i$ is the number of
tokens available after the match. Verification returns the accept length $A_i$, excluding tokens generated directly by the target model.
As some retrieval engines, such as SuffixDecoding \citep{suffixdecoding2025}, have their own draft length control policy, \sys uses the smaller value for the draft length between our policy and theirs.

\textbf{Offline profiling of $K_{\max}$.}\;
\sys profiles the cap from the trajectories already collected to build
$\mathcal{G}$. It replays them and records, at each retrieval step, the reusable
length $R$: the number of retrieved tokens that match the recorded continuation.
The distribution of $R$ for each agent gives the probability
$\Pr(R \ge k \mid M,a)$ that the $k$-th draft token would be accepted, which
falls as $k$ grows. A draft token pays off only if this probability exceeds
$\lambda$, the acceptance probability at which one more draft position covers
its verification cost (\Cref{sec:app-impl}). The cap is therefore the last position
that still pays off:
\begin{equation}
K_{\max}^{(M,a)} = \max\big\{k : \Pr(R \ge k \mid M,a) \ge \lambda\big\}.
\label{eq:design-cap}
\end{equation}
For example, suppose $\lambda = 0.1$, so that verifying one more position costs
as much time as producing $0.1$ tokens. If an agent's 11th draft token is
accepted with probability $0.12$ and its 12th with $0.09$, the 11th returns
$0.12$ tokens in expectation for a cost of $0.1$, whereas the 12th returns only
$0.09$; the cap is therefore 11.

\textbf{Online adaptation of $\alpha$.}\;
Within a session, the useful length for the same agent drifts across and within turns. \sys therefore
adapts each agent's scale from verification feedback. Each scale starts at
$1.0$, which requests exactly as many tokens as were matched, and after step $i$
is updated as
\begin{equation}
\alpha^{(a_i)} \leftarrow \alpha^{(a_i)}
  + \eta\big(\tau - \mathbb{I}[A_i < D_i]\big),
\label{eq:design-update}
\end{equation}
where $\mathbb{I}[\cdot]$ is the indicator function and $\eta$ is the step size.
Under the idealization in \Cref{app:alpha-update}, \Cref{eq:design-update} is a
stochastic subgradient step on the quantile loss
$\rho_\tau(R_i/m_i - \alpha^{(a_i)})$ \citep{koenker1978regression}, where $R_i$
is the number of tokens the target would accept from an unbounded draft.
The scale thus tracks the $\tau$-quantile of $R_i/m_i$, targeting full
acceptance ($A_i = D_i$) with probability $1-\tau$; setting $\tau = 1-\lambda$
aligns this target with the cap's break-even point, now for the current session
rather than the agent's history.
We skip the update when a fully accepted draft reaches
$K_{\max}^{(M,a_i)}$, since the cap censors the observation. The scales persist across sessions served by the same server and are initialized once at server start.

Together, the cap and the scale apply one criterion, drafting a token only if it
is likely enough to be accepted to cover its verification cost: the cap from
each agent's past trajectories, and the scale from recent
verification feedback.
\section{Evaluation}
\label{sec:eval}

\subsection{Experimental Setup}
\label{sec:eval-setup}

\noindent\textbf{Models.}
We evaluate Devstral-24B \citep{devstral2025mistral}, Gemma3-27B \citep{gemma3},
and Qwen3.6-27B \citep{qwen3.6}.

\noindent\textbf{Benchmarks.}
Our main evaluation uses two repository-level benchmarks, SWE-bench Verified
\citep{jimenez2024swebench} and TeamBench \citep{teambench2026}.
On SWE-bench Verified, five agents---localizer, coder, executor, reviewer, and
summarizer---run in containers for up to five turns, with the coder editing
repository files through unified diffs.
TeamBench provides each task with a repository workspace and spans eighteen task
families, including code review, incident response, and infrastructure changes.
A planner, executor, and verifier proceed through planning, execution, and
verification, with up to 20 turns per phase and two remediation loops, capped
at 30 LLM calls per session.
We additionally evaluate \sys on LiveCodeBench \citep{jain2025livecodebench}
and Terminal-Bench \citep{merrill2026terminalbench} to separately test
generality: LiveCodeBench has no repository, while Terminal-Bench uses a single
agent for all roles (\Cref{sec:eval-generality}).
We measure generation throughput excluding prefill and accept length (AL) for
all benchmarks.
% at batch size 16, throughput is measured over steady-state
% decoding steps with an active batch of at least 8 requests.

\noindent\textbf{Baselines.}
We compare autoregressive (AR) decoding, PLD \citep{saxena2023prompt}, REST
\citep{he2024rest}, FastCoder \citep{zhao2025fastcoder}, SuffixDecoding
\citep{suffixdecoding2025}, SAM-Decoding \citep{samdecoding2025}, and EAGLE-3
\citep{li2025eagle3}. We follow the draft-length settings in the original papers, with configuration
and implementation details reported in \Cref{sec:app-impl}. FastCoder extends REST with a
project datastore, a cache of verified drafts and model output, and a rule that
skips retrieval where it rarely succeeds; we attach its project datastore on
both SWE-bench Verified and TeamBench.

\noindent\textbf{\sys Implementation.}
We implement \sys in vLLM \citep{kwon2023vllm} over two retrieval engines: the
suffix automaton of SAM-Decoding (\sys-SAM) \citep{samdecoding2025} and the suffix tree of SuffixDecoding
(\sys-Suffix) \citep{suffixdecoding2025}.
Both engines use the same corpus configuration, draft-length controller,
per-agent caps, and workspace scope, differing only in the underlying retrieval
structure. For \sys-SAM, we use frequency-majority continuation selection at
both batch sizes. The maximum draft length is 64 at batch size 1 and 24 at batch size 16
for both implementations.
Detailed configuration is at \Cref{sec:app-impl}.

\noindent\textbf{Corpus construction.}
For each benchmark, we construct the global corpus from agent turn logs
collected on a disjoint set of coding tasks before evaluation.
We use 70 tasks from SWE-bench Verified and 213 from TeamBench,
indexing generated text and tool calls rendered in each model's native format.
Evaluation tasks are held out from corpus construction.

\subsection{Main Evaluation}
\label{sec:eval-main}
\label{sec:eval-prior}

% Main evaluation: 2 benchmarks x 3 models, 4 metric/batch rows (STORYLINE 9-AJ).
% Each axis will compare methods as in data/5_main.pdf. No results are imputed.
\begin{figure}[t!]
\centering
\includegraphics[width=\linewidth]{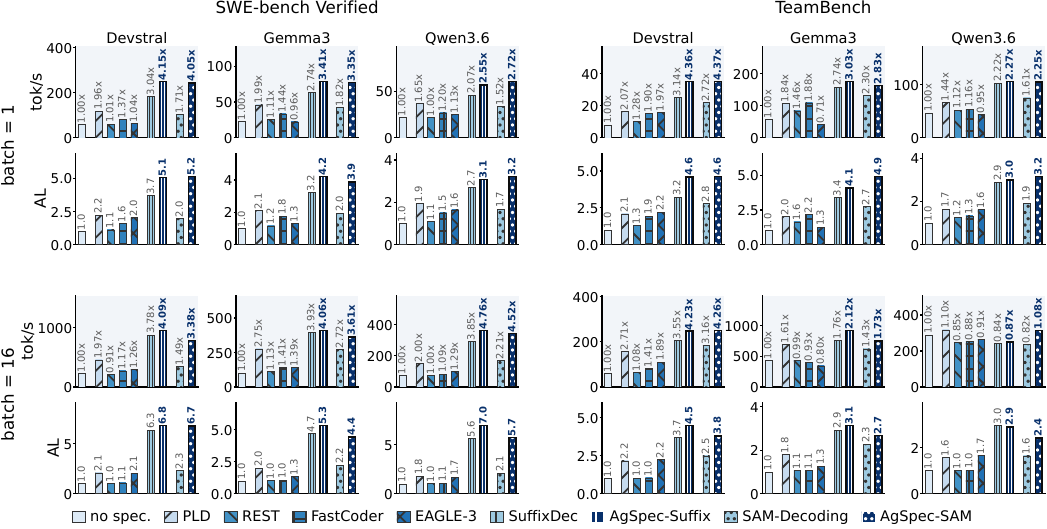}
\caption{Main comparison across two repository benchmarks and three models.
Rows show generation throughput and AL at batch sizes $1$ and $16$;
higher is better. The number above each bar is its ratio to the
no-speculation bar in that panel. no spec. denotes AR generation.}
\label{fig:main}
\label{fig:eval-prior}
\end{figure}

% TODO(exp): Confirm the second repository benchmark, cohort and repository
% coverage. Supply all 2 x 3 x 2 benchmark/model/batch configurations with
% native baselines and the current AgSpec corpus/controller configuration.
% Preserve native retrieval/adaptation and document method versions,
% target/tokenizer, serving conditions, corpus preparation and chronology.
% Match method/no-spec cohorts; report repetitions, uncertainty, trajectory
% pairing, retained coverage and achieved concurrency, especially at batch 16.
% Do not substitute AgSpec ablations or independently published numbers for
% native implementations. data/5_main.tsv remains earlier-configuration evidence.

\Cref{fig:main} compares \sys with autoregressive decoding and six prior
speculative methods on SWE-bench Verified and TeamBench, using three models at
batch sizes $1$ and $16$. Within each panel, all methods share the same
workload and serving configuration.

\sys achieves the highest or the second highest throughput in all evaluated settings,
reaching 2.27--4.37$\times$ the throughput of autoregressive decoding
at batch size 1 and 1.08--4.76$\times$ at batch size 16.
On average, its throughput is 18.0\% higher than that
of the fastest prior method.
The gains are largest with Devstral at batch size 1 on SWE-bench Verified, achieving 1.36$\times$ higher than the fastest prior work.
\sys also achieves the highest AL in most evaluated settings.
This demonstrates the effectiveness of \sys for coding agents.

Because \sys runs on top of the retrieval engine, we also evaluate each
engine with and without it.
Both engines benefit in most settings.
With SuffixDecoding, throughput rises by 16.5\% on average
and AL by 18.5\%; with SAM-Decoding, throughput rises by 64.6\% and AL nearly doubles.
The gap reflects what each engine already provides.
SuffixDecoding keeps a tree of past responses and scales its draft with
the match length, whereas SAM-Decoding does neither.
Thus, although SAM-Decoding trails SuffixDecoding in all twelve settings,
\sys-SAM outperforms it in nine.
These results show that existing retrieval engines, on their own,
leave much of the speedup available in coding-agent pipelines untapped
and \sys, as a framework, accelerates existing retrieval engines in
coding-agent pipelines without modification.

% Applying the \sys framework to existing retrieval engines yields higher
% generation throughput and accept length than the original
% engines, autoregressive decoding, and prior speculative methods overall. In each setting,
% the \sys variant achieves $2.27$--$4.37\times$ the throughput of
% autoregressive decoding at batch size $1$ and $1.24$--$4.76\times$ at batch
% size $16$. Compared with the fastest prior speculative method in each setting,
% its throughput improves by $19\%$ on average and up to $39\%$. Its AL is
% $18\%$ higher on average and up to $44\%$ higher than that of the prior
% speculative method with the highest AL, and $35\%$ higher on average and up to
% $88\%$ higher than that of its corresponding retrieval-engine baseline.
% Although baseline AL and the size of the gains vary across models and
% benchmarks, \sys captures reuse in session and workspace text and adjusts
% draft length, improving throughput with both engines and AL in most settings.

% TODO(exp): Replace the \sys rows once results under the current three-corpus
% and contextual-profile configuration are available.
% TODO(check): Reconcile the \sys run reported in data/5_main.tsv with the run
% recorded in Appendix~\ref{sec:app-result-details} before finalizing the results.

\subsection{Non-Repository and Implicit-Agent Tests}
\label{sec:eval-generality}

%LiveCodeBench tests generation without a repository. A coder and feedback agent
%iterate for up to five turns; one problem and its execution environment form a
%session. \sys uses the global reference and session corpora only: \sys uses the global reference and session corpora only: there is no repository
%workspace from which to open artifacts, so the workspace corpus stays empty. This tests behavior when the workspace component has no content
%(\Cref{fig:eval-generality}).

%Terminal-Bench tests implicit agent roles: a single Terminus agent performs
%functions such as inspecting files, taking actions, and interpreting feedback
%without separate role assignments. Corpus retrieval operates on the session
%regardless of these functions, while the agent-indexed length profile uses the
%defined implicit-agent mapping or fallback (\Cref{fig:eval-generality}).

\begin{figure}[t]
\centering
\includegraphics[width=\linewidth]{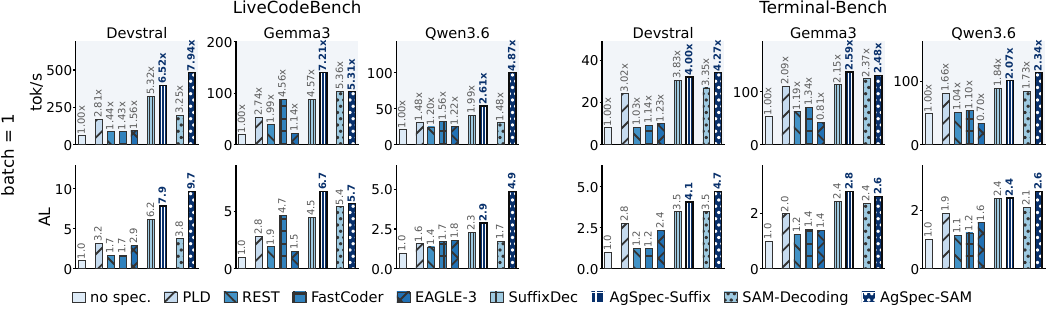}
\caption{Non-repository and single-agent evaluation at batch size $1$ across
three models. LiveCodeBench evaluates \sys without repository context, using the
global reference and session corpora with an empty workspace corpus.
Terminal-Bench evaluates \sys when agent functions are not represented by
separate explicit roles.}
\label{fig:eval-generality}
\end{figure}

\sys primarily targets repository-level coding tasks performed by multi-agent
systems. In practice, however, coding workloads also include simpler
scenarios, such as standalone code generation or single-agent pipelines like
ReAct~\citep{yao2023react}. To evaluate whether \sys generalizes to these
settings, \Cref{fig:eval-generality} shows its performance without a
repository workspace or explicit agent roles.

On LiveCodeBench, a coder and feedback agent work on each problem for
up to five turns without a repository.
Across the three models, the faster \sys variant reaches
4.87--7.94$\times$ the generation throughput of autoregressive decoding
and exceeds the fastest prior speculative method by 35--145\%.
Its accept length also exceeds that of the prior method with the
highest accept length.
Although the workspace corpus is unavailable, \sys can still retrieve
from the global corpus and text accumulated within the session,
while its draft-length controller continues to adapt from verification
feedback, yielding these gains.

On Terminal-Bench, a single Terminus agent inspects files, takes actions,
and interprets feedback without separate role assignments.
The faster \sys variant improves throughput over the fastest prior
speculative method by 9.3--27.2\%, while its accept length is
16.7--34.3\% higher than that of its corresponding retrieval-engine baseline.
For draft-length control, \sys uses a shared cap and a single $\alpha$
table instead of role-specific caps, updating the scale from verification feedback.
Thus, \sys can improve decoding even when the workflow has only one
explicit agent.
Together, these results show that \sys applies to a broader range of coding agents.
% Although roles are not
% explicitly separated, boundaries between task stages and tool calls
% remain in the session text and provide context for retrieval. Opened
% files are also reused through the workspace corpus. 

% TODO(exp): Run both benchmarks across all three models with native baselines
% and the current \sys configuration; report matched cohorts, profile mapping,
% repeats, and uncertainty.

\subsection{Analysis}

\subsubsection{Impact of Coding-Agent-Aware Retrieval}
\label{sec:eval-ladder}

\begin{figure}[htbp]
\centering
\includegraphics[width=\linewidth]{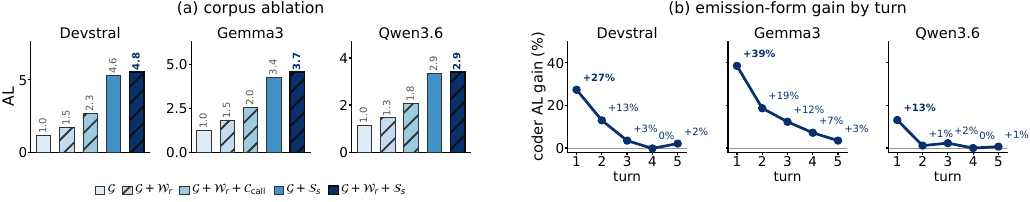}
\caption{Corpus ablation on SWE-bench Verified at batch size $1$.
(a)~AL as corpora are added; the number above each bar is its ratio to
$\mathcal{G}$ alone, and $\mathcal{C}_{\mathrm{call}}$ is a per-call corpus of
the current prompt and output. (b)~Coder AL gain per turn from indexing
workspace artifacts in emission form, relative to raw file text only.}
\label{fig:eval-corpus}
\end{figure}

We measure the contribution of coding-agent-aware retrieval through
trace-driven simulation. In particular, we examine two design choices:
using three corpora from different sources and indexing artifacts based on emission format.
We replay the drafter over traces of \sys-SAM sessions
on SWE-bench Verified under each corpus configuration and compare the expected AL (\Cref{fig:eval-corpus}).

\Cref{fig:eval-corpus}(a) adds corpora to the global corpus one at a time.
Retaining text across calls gives the largest gain: replacing a per-call
corpus with the session corpus raises AL by $1.6$--$2.1\times$, as agents
repeatedly reproduce code and outputs from earlier calls. The workspace corpus
adds $26$--$47\%$ on its own and still $2.2$--$7.0\%$ on top of the session
corpus, bringing \sys to $2.9$--$4.8\times$ the AL of the global corpus alone. This shows that each source contributes significantly to the system.

\Cref{fig:eval-corpus}(b) shows that emission-form indexing helps most in the
first turn, raising the coder's AL by $13$--$39\%$ before its own patches enter
the session corpus. From the second turn on, those patches already supply the
emitted form, so the gain falls to $3\%$ or less by the fifth turn. Together,
these results show that the session corpus supplies most of the reuse, while
the workspace corpus and its emission forms cover what the session has not yet
accumulated.

\subsubsection{Impact of Agent-Aware Draft-Length Control}
\label{sec:adaptive-length}

\begin{figure}[thbp]
\centering
\includegraphics[width=0.99\linewidth]{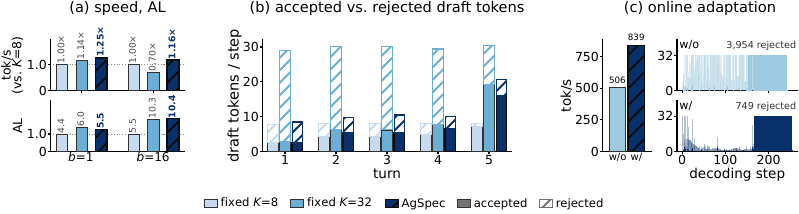}
\caption{Draft-length control on SWE-bench Verified with Devstral. (a)~Generation throughput
(top) and AL (bottom) of fixed draft lengths $K$ (every agent capped at $K$,
no online scale) and \sys, normalized to $K=8$
within each group. (b)~Mean draft tokens per drafting step by turn at batch
size $1$, split into accepted (solid) and rejected (hatched) tokens.
(c)~Fixed $K=32$ without and with the online scale enabled, at batch
size $16$: throughput (left) and case study of coder call per decoding step
(right), with accepted tokens solid and rejected tokens lightly shaded.}
\label{fig:eval-length}
\end{figure}

We measure the contribution of agent-aware draft-length control on SWE-bench
Verified with Devstral (\Cref{fig:eval-length}). We use SAM-Decoding as the engine since its draft length is a static cap,
which lets the comparison reflect our control alone.  Within \sys-SAM, we compare our control against fixed caps, $K \in \{8,32\}$, and separately ablate the online scale.

\Cref{fig:eval-length}(a) shows that the benefit of long drafts disappears at
large batch sizes. At batch size 1, $K=32$ is 14\% faster than $K=8$, but at batch size 16
it is 30\% slower, as rejected tokens consume compute shared by the whole batch.
\sys is the fastest at both batch sizes, 25\% and 16\% above $K=8$.
$K=32$ attains a higher AL than \sys at batch size 1 (6.0 vs.\ 5.5)
and a slightly lower AL at batch size 16 (10.3 vs.\ 10.4),
but remains slower at both batch sizes because of the drafts it rejects.

\Cref{fig:eval-length}(b) explains this at batch size $1$. Rejected tokens add verification cost without producing output,
so throughput depends on accepting many tokens while rejecting few.
$K=32$ rejects 11--26 tokens per step, whereas \sys rejects at most 6,
close to $K=8$, while still accepting 83--89\% as many tokens as $K=32$.
\sys lengthens its drafts only as reusable text accumulates, from about
10 tokens in early turns to 21 in the fifth, and thus achieves the
highest throughput.

\Cref{fig:eval-length}(c) isolates the online scale at batch size $16$. To measure its effect alone, we enable only the online scale on top of
the fixed $K=32$ draft length of Fig.~7(a), without per-agent caps.
The online scale alone raises throughput by 66\%, as the share of
rejected draft tokens falls from 78\% to 50\%.
The example coder call shows how: without it, the drafter spends about
26 tokens per step over the first 160 steps but has fewer than 2 accepted;
with it, the drafter shortens its drafts to about 4 tokens there and
expands to the full draft length once the coder starts copying reusable text.
It thus rejects 81.1\% fewer tokens while taking only 11 more steps.

% SOURCE: AgSpec.code data/09_ablations/5_length_live.tsv <- benchmarks/export_length_live.py
% (live runs: results/analysis_live/{k8,k32,b16_k8,b16_k32}/devstral_24b, results/ours_atab/devstral_24b/swe_b{1,16}).
% b16 tok/s = pooled (batch_summary); w/ vs w/o online adaptation pending a live run.

% \input{tex/6Discussion}  % removed from the main text 2026-09-24
\section{Conclusion}
\label{sec:conclusion}

We proposed \sys, a retrieval-based SD framework for coding-agent pipelines.
\sys has two key designs: it makes the text an agent reuses available
for retrieval in the form the agent emits, and it adapts the draft
length to each agent using profiled accept lengths and verification feedback.
Because \sys is orthogonal to the retrieval engine, it can run on top
of existing engines without modification.
Our evaluation shows that \sys achieves the highest throughput in most
evaluated settings on SWE-bench Verified and TeamBench, reaching
2.27--4.37$\times$ the throughput of autoregressive decoding at batch size 1,
and remains effective without a repository or explicit agent roles.
These results show that retrieval-based SD for coding agents should
follow the agent pipeline, as \sys does: it indexes the text the session has produced, in the form agents emit it, and drafts as much as each agent can reuse.

% Appendix~\ref{sec:eval-protocol} provides the serving configuration,
% corpus construction, baseline settings, evaluation protocol, and metric
% definitions. Appendices~\ref{sec:app-operations}--\ref{sec:app-history-size}
% provide additional ablations and analyses. The supplementary material
% includes experiment configurations, dataset splits, corpus-construction
% scripts, and evaluation scripts.
% ICLR 2027 AI policy (https://iclr.cc/Conferences/2027/AIPolicyForAuthors):
% mandatory section, not counted toward the page limit. Wording follows the
% policy's template. If any required-disclosure task used generative AI (e.g.
% implementing methods, designing experiments, interpreting results), add it here.
\section*{Use of Generative AI Tools}
In this work, we used generative AI tools to assist with translation, drafting and editing manuscript text, and providing feedback on figures and the presentation of our methods and experiments. 
Also, we exploited AI to get advice for the justification of our online adaptation mentioned at \Cref{app:alpha-update}. 
The authors reviewed AI-assisted suggestions and take responsibility for the final content of this work, including text, claims, and artifacts produced with the aid of generative AI.

\section*{Ethics statement}

\sys verifies drafts with the target model, preserving its output distribution for each model call. Its decoding method therefore does not introduce new output-level behavior relative to ordinary target-model decoding. The additional concern is that the workspace and session corpus may contain sensitive prompts, repository content, or tool outputs. Deployments should restrict access to each session's corpus, prevent its disclosure to other sessions, and discard it when the session ends.

\section*{Reproducibility Statement}

\Cref{sec:design-retrieval,sec:design-length} specify the retrieval and draft-length algorithms. \Cref{sec:eval-setup,sec:app-impl} report the task splits, corpus construction, and measurement protocol, while \Cref{tab:app-config} lists the model and serving configurations. For each model and benchmark, we built the global corpus from tasks disjoint from evaluation and held it fixed during measurement (\Cref{sec:eval-setup,sec:app-impl,tab:app-config}); session-specific corpora were reset between sessions (\Cref{tab:app-config}). We used greedy decoding (\Cref{tab:app-config}) and the same fixed seed across runs, including model calls that generate tool-call requests. Compared methods used the same evaluation tasks and serving configuration (\Cref{sec:eval-main}).

\bibliography{iclr2025_conference}
\bibliographystyle{iclr2025_conference}

\appendix
\newpage

\section{Implementation Details}
\label{sec:eval-protocol}
\label{sec:app-impl}

\begin{table}[!t]
\centering
\small
\caption{Configuration of the reported \sys experiments.}
\label{tab:app-config}
\begin{tabular}{|>{\raggedright\arraybackslash}p{0.20\linewidth}|
>{\raggedright\arraybackslash}p{0.72\linewidth}|}
\hline

Hardware &
One NVIDIA RTX PRO 6000 Blackwell Server Edition (96\,GB) per run. \\
\hline

Serving &
vLLM 0.17.1 (V1 engine), TP1, eager execution, prefix caching off,
chunked prefill on, and maximum model length $16{,}000$. \\
\hline

Models &
Devstral-Small-2507 AWQ 4-bit, Gemma3-27B-it W4A16, and
Qwen3.6-27B AWQ INT4, with greedy decoding. \\
\hline

Global corpus $\mathcal{G}$ &
One fixed corpus per model and benchmark, constructed before serving. \\
\hline

Session corpus $\mathcal{S}_s$ &
Trailing $2{,}048$ prompt tokens and generated tokens, shared within a
session and reset afterward. \\
\hline

Workspace corpus $\mathcal{W}_r$ &
Current-session repository artifacts in raw and emission forms, reset
after the session. \\
\hline

Retrieval &
Minimum match length $1$ and margin $\delta=5$ against $\mathcal{G}$.
$\mathcal{G}$ and $\mathcal{S}_s$ use frequency; $\mathcal{W}_r$ uses
the first occurrence. \\
\hline

Draft control &
$\alpha_0=1.0$, $\eta=0.5$, $\alpha\in[0.05,12.0]$, and
$\tau=1-\lambda$. Scales initialized once per server. \\
\hline

Draft-length limits &
Maximum draft length $64$ at batch size $1$ and $24$ at batch size $16$;
model- and agent-specific draft caps are profiled offline within these limits. \\
\hline

\end{tabular}
\end{table}

\paragraph{Baseline draft lengths.}
Table~\ref{tab:app-baseline} lists the draft length $K$ and the engine settings
of each baseline. REST and FastCoder draft over a token tree, so their
published policy has no direct chain-length equivalent; we keep the chain
approximation of our implementation and report it as such rather than mapping
the tree depth to a chain length. SuffixDecoding uses the default
configuration of the vLLM implementation. EAGLE-3 uses a chain of depth $3$,
matching the $3$ drafting steps of the published $3$-$4$-$8$ tree. The
autoregressive baseline runs through the same proposer path with the match
threshold raised so that no draft is produced, which keeps the per-step timing
logs identical across arms. Baseline settings were fixed before measurement and
were not revised afterward.

\paragraph{Additional \sys Implementation}
The session corpus admits the trailing $2{,}048$ tokens of each prompt and is
reset between sessions. The workspace corpus is likewise isolated by session,
so artifacts from other evaluation sessions are never exposed to retrieval.
The margin $\delta$ is $5$ tokens. For the draft length,
$\eta=0.5$; the scale is initialized to $1.0$, clipped to $[0.05,12.0]$,
and not reset between sessions.

\begin{table}[!t]
\centering
\small
\caption{Baseline draft lengths and engine settings.}
\label{tab:app-baseline}
\begin{tabular}{|l|c|>{\raggedright\arraybackslash}p{0.52\linewidth}|}
\hline
Method & $K$ & Settings \\
% \hline
% Autoregressive & 1 & Match threshold $99{,}999$; no draft produced \\
\hline
PLD & 10 & $n$-gram length $3$--$5$ over the prompt \\
\hline
REST & 10 & Match threshold $2$; per-model datastore \\
\hline
FastCoder & 10 & Match threshold $2$; same datastore as REST \\
\hline
SAM-Decoding & 15 & Match threshold $2$; $1+15$ verified tokens per step \\
\hline
SuffixDecoding & 24 & Tree depth $24$, speculation factor $1.0$, minimum token probability $0.1$ \\
\hline
EAGLE-3 & 3 & Chain depth $3$; per-model drafter checkpoint \\
\hline
\end{tabular}
\end{table}

\paragraph{Corpus construction.}
For each model and benchmark, $\mathcal{G}$ is built from instances disjoint
from evaluation. SWE-bench Verified uses $70$ corpus-building and $30$
evaluation instances; TeamBench uses $213$ and $30$, respectively.

\paragraph{Break-even threshold.}
We estimate
\[
\lambda=\bar{A}\frac{c}{S},
\]
where $S$ is target-model step time, $c=dS/dL$ is the marginal cost of one
additional draft token, and $\bar{A}$ is the mean accepted tokens per step.
We use $\lambda$ for the offline cap and set $\tau=1-\lambda$. The caps and
$\tau$ of the reported runs were fixed from a cost profile taken before the
final runs and were not revised afterward. \Cref{tab:app-tau-kmax} lists
$\lambda$ re-measured on the final batch-size-16 runs themselves, over steps
with at least 12 active requests, next to the $\tau$ and per-agent caps
actually used.

\begin{table}[htbp]
\centering
\small
\caption{Break-even threshold re-measured on the final batch-size-16 runs
(steps with $b \ge 12$), with the $\tau$ and per-agent caps $K_{\max}$ used
in the reported runs. Caps are listed in agent order: planner / executor /
verifier for TeamBench, and localizer / coder / executor / feedback /
summarizer for SWE-bench Verified; on SWE-bench Verified only the first three
are set explicitly and the other two default to the engine budget. On SWE-bench Verified, all three models
use the $\lambda$ measured with Devstral.}
\label{tab:app-tau-kmax}
\begin{tabular}{llrrrl}
\hline
Benchmark & Model & $\lambda$ & $1-\lambda$ & $\tau$ used & $K_{\max}$ used \\
\hline
TeamBench          & Devstral & $0.022$ & $0.979$ & $0.944$ & 4 / 14 / 19 \\
                   & Gemma3   & $0.133$ & $0.867$ & $0.944$ & 4 / 24 / 11 \\
                   & Qwen3.6  & $0.051$ & $0.949$ & $0.896$ & 15 / 3 / 7 \\
SWE-bench Verified & Devstral & $0.299$ & $0.701$ & $0.950$ & 24 / 24 / 24 / 24 / 24 \\
                   & Gemma3   & $0.299$ & $0.701$ & $0.950$ & 24 / 24 / 24 / 24 / 24 \\
                   & Qwen3.6  & $0.299$ & $0.701$ & $0.950$ & 24 / 24 / 24 / 24 / 24 \\
\hline
\end{tabular}
\end{table}

\section{Additional Corpus Ablations}
\label{sec:app-operations}

We further analyze three aspects of the corpus design: the sharing scope of
the session corpus, how the workspace corpus grows, and whether it is kept
separate from the global corpus (\Cref{tab:app-corpus-controls}).

\begin{table}[htbp]
\centering
\small
\caption{Additional corpus controls on SWE-bench Verified at batch size $1$.
Values are AL changes relative to the reported configuration. Asterisks mark a two-sided Wilcoxon signed-rank test over paired sessions ($^{*}p<0.05$, $^{**}p<0.01$, $^{***}p<0.001$; no multiple-comparison correction).}
\label{tab:app-corpus-controls}
\begin{tabular}{lccc}
\hline
Control & Devstral & Gemma3 & Qwen3.6 \\
\hline
Per-call session corpus
    & $-51.4\%^{***}$ & $-44.2\%^{***}$ & $-39.3\%^{***}$ \\
Per-agent session corpus
    & $-2.3\%^{***}$ & $-1.7\%^{***}$ & $-1.4\%^{***}$ \\
Cross-session workspace growth
    & $-0.2\%$ & $-0.5\%$ & $+0.3\%$ \\
Workspace pooled into $\mathcal{G}$
    & $+0.2\%$ & $0.0\%$ & $+0.5\%$ \\
\hline
\end{tabular}
\end{table}

The session corpus must retain trajectory context across calls: restricting
$\mathcal{S}_s$ to the current call sharply reduces AL, because coding agents
frequently reuse content from earlier inputs and outputs. Sharing
$\mathcal{S}_s$ across agents yields a further, smaller gain, as agents also
reuse what other agents have read or produced. 

Letting $\mathcal{W}_r$ grow
across sessions brings no meaningful benefit, as artifacts from the current
session already capture most of the useful workspace reuse. This may change
when many sessions run on the same repository, which we leave for future
work. 

Finally, pooling $\mathcal{W}_r$ into $\mathcal{G}$ performs on par with
keeping it separate. We nevertheless keep them separate: in real serving
environments, each user works on a different repository, and folding every
workspace into a global corpus shared by all users would be difficult to
manage.

\section{Corpus Preference Margin}
\label{sec:app-pref-margin}

We vary the preference margin $\delta$ to test the sensitivity of corpus
selection to the bias against $\mathcal{G}$.

\begin{table}[htbp]
\centering
\small
\caption{Preference-margin sensitivity on SWE-bench Verified ($b=1$).
Values are oracle-normalized continuation agreement (\%).}
\label{tab:app-pref-margin}
\begin{tabular}{lccc}
\hline
$\delta$ & Devstral & Gemma3 & Qwen3.6 \\
\hline
$0$         & $93.8^{**}$  & $93.3^{***}$ & $90.8^{***}$ \\
$2$         & $95.0^{**}$  & $94.7^{***}$ & $93.6^{**}$  \\
$5^\dagger$ & $94.6$       & $94.1$       & $93.3$       \\
$8$         & $94.3^{***}$ & $93.8^{***}$ & $93.2^{***}$ \\
$12$        & $94.2^{***}$ & $93.7^{***}$ & $93.2^{***}$ \\
$\infty$    & $94.1^{***}$ & $93.6^{***}$ & $93.2^{***}$ \\
\hline
\end{tabular}
\end{table}

Agreement remains high across a broad range of margins, with $\delta=2$
giving the highest values across all three models. The reported $\delta=5$
therefore does not rely on a sharply tuned preference margin.

\section{Corpus-Specific Continuation Selection}
\label{sec:appendix-selection}

Because the corpora capture different forms of reuse, we replay the same
matches and candidates under alternative continuation-selection rules.

\begin{table}[htbp]
\centering
\small
\caption{Continuation-selection replay on SWE-bench Verified. Values are
changes in reference-matching continuation length (\%) relative to the
reported rule, averaged across three models.}
\label{tab:app-selection}
\begin{tabular}{lccc}
\hline
Rule & Global $\mathcal{G}$ & Session $\mathcal{S}_s$ & Workspace $\mathcal{W}_r$ \\
\hline
first occurrence
    & $-18.1^{***}$ & $-4.2^{***}$ & $0^\dagger$ \\
Most recent
    & $-7.9^{***}$ & $+1.1^{***}$ & $-2.6^{**}$ \\
Most frequent
    & $0^\dagger$ & $0^\dagger$ & $-1.0$ \\
REST-style trie
    & $+6.5^{***}$ & $+0.1$ & $+0.3$ \\
Recency--frequency
    & $-0.8^{***}$ & $+0.6^{***}$ & $-1.1$ \\
\hline
\end{tabular}

\vspace{2pt}
\footnotesize
$\dagger$: reported rule.
$^{*}p<0.05$, $^{**}p<0.01$, $^{***}p<0.001$
(two-sided Wilcoxon signed-rank test over paired sessions; uncorrected).
\end{table}

The corpora favor different selection behavior: $\mathcal{G}$ benefits from
frequency, while $\mathcal{W}_r$ favors its earliest occurrence.
For $\mathcal{S}_s$, alternative rules yield only small differences, so we
reuse the frequency rule of $\mathcal{G}$ rather than introducing a separate
session-specific rule. REST-style selection shows additional headroom for
structured candidate ranking in $\mathcal{G}$.

\section{Draft Token Composition}
\label{sec:app-draft-mix}

\begin{figure}[!t]
\centering
\includegraphics[width=\linewidth]{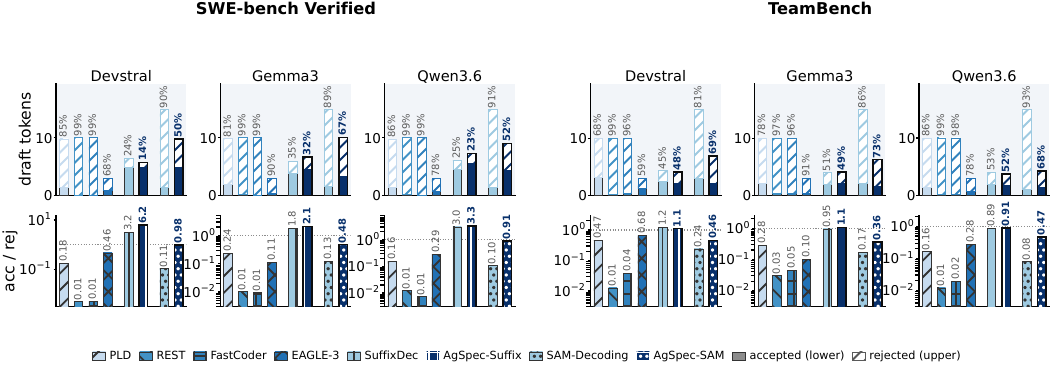}
\caption{Draft-token composition at batch size $16$ over decoding steps
with at least $12$ active requests. Top: mean accepted (solid) and
rejected (hatched) draft tokens per drafting step; labels report rejection
rates. Bottom: the ratio of accepted to rejected draft tokens, on a log
scale, with the dotted line at equal counts. No-speculation is omitted.}
\label{fig:app-draft-mix}
\end{figure}

\Cref{fig:app-draft-mix} shows the average accepted and rejected token length at batch size 16.
On the suffix automaton, \sys reduces rejection rates over SAM-Decoding by
$12.0$--$39.8$ points while drafting on $98$--$99\%$ of request steps,
compared with $50$--$56\%$ for SAM-Decoding. This shows that the reduction
comes from shorter, better-placed drafts rather than less frequent drafting.
On the suffix tree, rejection rates are similar on TeamBench, while
\sys-Suffix reduces them by $1.9$--$10.0$ points on SWE-bench Verified.

\begin{figure}[tbp]
\centering
\includegraphics[width=\linewidth]{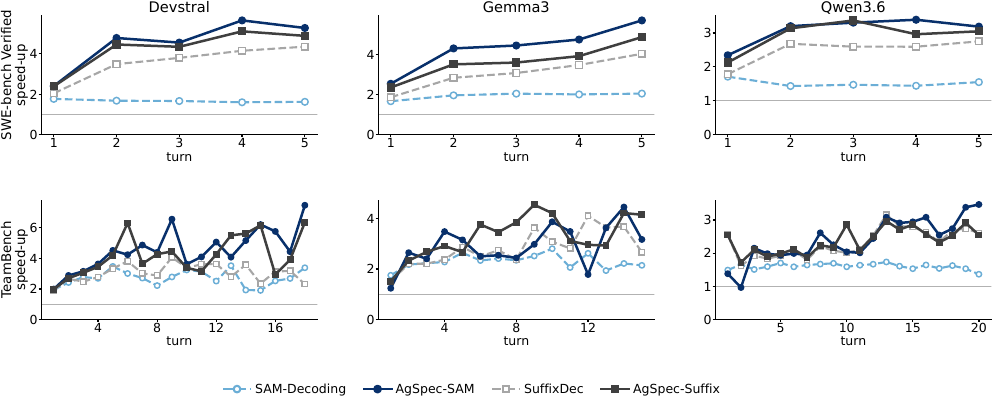}
\caption{Speed-up over no speculation by turn at batch size $1$ on
SWE-bench Verified and TeamBench.}
\label{fig:app-turn-speedup}
\end{figure}

\section{Speed-up by Turn and Agent}
\label{sec:app-turn-agent}

\begin{figure}[tbp]
\centering
\includegraphics[width=\linewidth]{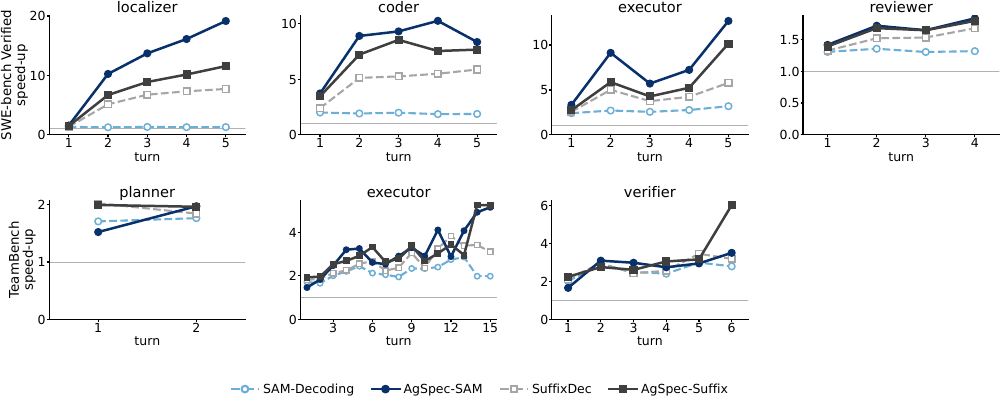}
\caption{Speed-up over no speculation by agent and turn at batch size $1$.
TeamBench results are geometric means across available models.}
\label{fig:app-agent-speedup}
\end{figure}

We examine whether speculative-decoding gains remain uniform across agents (\Cref{fig:app-agent-speedup})
and turns (\Cref{fig:app-turn-speedup}).

Speed-up varies substantially across both agents and trajectory positions,
and its turn-wise trend is not uniformly monotonic. This heterogeneity
supports adapting speculation to the observed trajectory rather than using
a fixed agent- or turn-level schedule.

\section{Discussion}
\label{sec:app-discussion}

\subsection{Interpreting the Scale Update}
\label{app:alpha-update}

Fix an agent and drop the superscript $(a_i)$. We idealize the draft length as
$D_i = \alpha m_i$, ignoring rounding, clipping of $\alpha$, the cap
$K_{\max}$, and corpus continuations shorter than $\alpha m_i$. Let $R_i$ be the
number of tokens the target would accept from an unbounded draft, so that
$A_i = \min(R_i, D_i)$ and
\begin{equation*}
\mathbb{I}[A_i < D_i] = \mathbb{I}[R_i < \alpha m_i] = \mathbb{I}[U_i < \alpha],
\qquad U_i = R_i / m_i .
\end{equation*}

\textbf{Objective.}\;
With the pinball loss $\rho_\tau(u) = u\,(\tau - \mathbb{I}[u < 0])$,
\begin{equation*}
\frac{\partial}{\partial \alpha}\,\rho_\tau(U_i - \alpha)
  = -\big(\tau - \mathbb{I}[U_i < \alpha]\big)
\end{equation*}
for $U_i \neq \alpha$, so \Cref{eq:design-update} is a stochastic subgradient
step on $\mathbb{E}[\rho_\tau(U - \alpha)]$, which is minimized at the
$\tau$-quantile of $U$. The loss is defined on the normalized length $U_i$
because $\alpha$ scales the draft relative to the match length. Applying it to
the raw length instead, $\rho_\tau(R_i - \alpha m_i)$, yields the subgradient
$-m_i(\tau - \mathbb{I}[R_i < \alpha m_i])$, which weights each step by $m_i$
and targets an $m$-weighted quantile of $U$; \Cref{eq:design-update} does not
apply this weighting.

\textbf{Equilibrium.}\;
If the distribution of $U_i$ were stationary, the expected update
$\eta\,(\tau - \Pr[U < \alpha])$ would vanish exactly when
$\Pr[U < \alpha] = \tau$, i.e., when drafts are fully accepted
($A_i = D_i$) with probability $1-\tau$. In practice, $\eta$ is fixed, sessions
are finite, and the distribution of $U_i$ shifts as $\mathcal{S}_s$ grows, so
$\alpha$ does not converge; it fluctuates around and tracks this moving target.
The ignored effects further perturb this balance. In particular, a fully
accepted draft truncated by $K_{\max}$ reveals only $R_i \ge K_{\max}$, not
whether $U_i \ge \alpha$; counting it as a full acceptance would bias $\alpha$
upward, which is why \sys skips such updates.

\subsection{Adaptive Draft Length Beyond Retrieval}
\label{sec:app-adaptive-related}

Adaptive draft length has been studied mainly for model-based SD, where a
draft model or head generates each token and exposes signals about how likely
that token is to be accepted. SpecDec++ trains an acceptance-prediction head
and stops drafting once the predicted rejection probability exceeds a
threshold \citep{huang2025specdecpp}. DISCO feeds the draft model's top-$k$
probabilities and entropy to a small classifier that decides whether to
continue \citep{mamou2024disco}, and AdaEDL and SVIP stop when the entropy of
the draft distribution signals a low acceptance probability
\citep{agrawal2024adaedl,zhang2025svip}. EAGLE-2 likewise expands its draft
tree along the draft model's most confident branches \citep{li2024eagle2}.

Retrieval-based SD has none of these signals. Its drafter copies a
continuation from a corpus instead of sampling one, so there is no token
distribution, entropy, or hidden state to read, and training a head to
predict acceptance would reintroduce the model that retrieval-based SD avoids.
What a retrieval drafter does have is the match itself, such as its length,
occurrence count, and source corpus, and the outcome of verification.
Existing retrieval methods either fix the draft length or scale it with the
match length (\Cref{sec:prelim}). SmartSpec uses verification outcomes,
choosing the length that maximizes estimated goodput under the observed
acceptance and system load \citep{liu2024smartspec}, but it targets general
request streams rather than agent pipelines, where acceptance differs
systematically across agents. \sys uses both available signals and adds the
one that coding agents provide, the agent issuing each call: an offline cap
per agent bounds the draft, and an online scale fitted to verification
feedback sizes it from the match length.

\subsection{Overhead of \sys}
\label{sec:app-overhead}

\paragraph{Memory.}
The global corpus $\mathcal{G}$ is loaded once per server and shared across
sessions. Its resident host-memory footprint is $2.5$--$4.3$\,GB, with
$9$--$15$\,s load time (\Cref{tab:app-overhead-mem}); it does not consume GPU
memory or reduce the KV cache. The session and workspace corpora are
session-local and released afterward.

\begin{table}[htbp]
\centering
\small
\caption{Global-corpus memory footprint and load time. Token counts are
reported for TeamBench.}
\label{tab:app-overhead-mem}
\begin{tabular}{llrrrr}
\hline
Benchmark & Model & Resident & On disk & Load & Tokens \\
\hline
SWE-bench Verified & Devstral & $3{,}603$\,MB & $214$\,MB & $12.4$\,s & -- \\
                   & Gemma3   & $4{,}121$\,MB & $253$\,MB & $14.1$\,s & -- \\
                   & Qwen3.6  & $4{,}295$\,MB & $258$\,MB & $14.7$\,s & -- \\
TeamBench          & Devstral & $2{,}484$\,MB & $144$\,MB & $9.2$\,s  & $1.94$\,M \\
                   & Gemma3   & $3{,}180$\,MB & $189$\,MB & $11.5$\,s & $2.51$\,M \\
                   & Qwen3.6  & $3{,}640$\,MB & $214$\,MB & $14.0$\,s & $2.94$\,M \\
\hline
\end{tabular}
\end{table}

\paragraph{Construction.}
Building $\mathcal{G}$ from recorded trajectories takes $18$--$26$\,s on CPU
and requires no GPU. $\mathcal{W}_r$ is built per session, while
$\mathcal{S}_s$ grows during decoding; both costs are included in reported
throughput.

\paragraph{Retrieval.}

At batch size $16$, querying the three corpora, selecting a draft, and
appending the step's tokens to the session corpus takes
$2.3$--$5.8$\,ms per step for \sys-SAM and $1.6$--$2.8$\,ms for \sys-Suffix.
A step takes 47--182 ms, so \sys accounts for 1.1--5.5\% of it
and the rest is the target model and the engine. We do not split the rest further:
the engine launches the forward pass asynchronously, so its CPU-side timer
covers only $2$--$56\%$ of the step.
\Cref{tab:app-overhead-step} gives the split per model.

\begin{table}[htbp]
\centering
\small
\caption{Per-step split at batch size $16$, over steps with at least $12$
active requests. \emph{\sys} is the proposer: corpus queries, draft selection,
and appending the step's tokens. \emph{Rest} is the target model and the
engine.}
\label{tab:app-overhead-step}
\begin{tabular}{llrrrr}
\hline
Benchmark & Model & Engine & \sys & Rest & \sys\ share \\
\hline
SWE-bench Verified & Devstral & SAM    & $5.05$\,ms & $98.1$\,ms  & $4.9\%$ \\
                   &          & Suffix & $2.80$\,ms & $78.7$\,ms  & $3.4\%$ \\
                   & Gemma3   & SAM    & $5.81$\,ms & $132.8$\,ms & $4.2\%$ \\
                   &          & Suffix & $2.57$\,ms & $142.6$\,ms & $1.8\%$ \\
                   & Qwen3.6  & SAM    & $2.73$\,ms & $162.6$\,ms & $1.7\%$ \\
                   &          & Suffix & $2.54$\,ms & $179.4$\,ms & $1.4\%$ \\
TeamBench          & Devstral & SAM    & $2.96$\,ms & $147.6$\,ms & $2.0\%$ \\
                   &          & Suffix & $1.57$\,ms & $145.6$\,ms & $1.1\%$ \\
                   & Gemma3   & SAM    & $2.77$\,ms & $47.9$\,ms  & $5.5\%$ \\
                   &          & Suffix & $1.60$\,ms & $45.8$\,ms  & $3.4\%$ \\
                   & Qwen3.6  & SAM    & $2.27$\,ms & $110.5$\,ms & $2.0\%$ \\
                   &          & Suffix & $1.94$\,ms & $165.1$\,ms & $1.2\%$ \\
\hline
\end{tabular}
\end{table}

\end{document}